# *ArrayTac*: A Closed-Loop Piezoelectric Tactile Platform for Continuously Tunable Rendering of Shape, Stiffness, and Friction

Tianhai Liang[1], Shiyi Guo[1,3], Baiye Cheng[1,4], Zhengrong Xue[1], Han Zhang[1], Huazhe Xu[1,2,3*]

[1] Institute for Interdisciplinary Information Sciences, Tsinghua University; Beijing, China.
[2] Shanghai Qi Zhi Institute; Shanghai, China.
[3] Pokebot Inc.; Beijing, China.
[4] School of Electronic Information and Communications, Huazhong University of Science and Technology; Wuhan, China.
*Corresponding author. Email: huazhe_xu@mail.tsinghua.edu.cn

## ABSTRACT

Human touch depends on the integration of shape, stiffness, and friction, yet existing tactile displays cannot render these cues together as continuously tunable, high-fidelity signals for intuitive perception. We present ArrayTac, a closed-loop piezoelectric tactile platform that simultaneously renders these three tactile dimensions with continuous tunability on a 4 by 4 actuator array. Each unit integrates a three-stage micro-lever amplifier with end-effector Hall-effect feedback, enabling up to 5 mm displacement, an array refresh rate greater than 500 Hz, and a closed-loop bandwidth of 123 Hz. In psychophysical experiments, naive participants identified three-dimensional shapes and distinguished multiple stiffness and friction levels through touch alone without training. We further demonstrate image-to-touch rendering from an RGB image and remote palpation of a medical-grade breast tumor phantom across a distance of more than 1,000 km, in which all 11 participants correctly identified the number and type of tumors with a sub-centimeter mean localization error. Together, these results establish ArrayTac as a programmable multidimensional tactile rendering platform that bridges digital representations, physical environments, and human tactile perception, enabling intuitive tactile communication between humans and machines.

## INTRODUCTION

Haptic perception is central to how humans interact with the physical world. Through the coordinated action of mechanoreceptors embedded in the skin, humans can discern an object's shape, stiffness, and friction[1–4]. This multidimensional tactile capability is essential not only for everyday manipulation and material identification, but also for tasks that require precise and intuitive feedback, such as medical palpation, precision manufacturing, and remote operation[5–7]. Although visual and auditory interfaces for virtual reality and teleoperation have matured considerably, haptic feedback remains a major bottleneck for achieving realistic human–machine interaction[1,8]. A key unresolved challenge is to render shape, stiffness, and friction simultaneously on a tactile display with sufficient fidelity and continuous tunability for practical use. The lack of this capability limits the realism, interpretability, and utility of current haptic systems[1,5,7–8]. Consequently, physical tactile experiences remain difficult to transmit between humans, machines, and digital environments, limiting the development of intuitive tactile communication and richer forms of human–machine interaction.

For realistic active touch, a tactile display should render shape, stiffness, and friction at high refresh rates. These cues jointly determine how humans perceive and interpret object surfaces[9]. To pursue this goal, prior work has explored diverse actuation strategies, including mechanically driven systems[10–12], electromagnetic devices[13–16], ultrasonic approaches[17–18], electrical stimulation interfaces[19–22], smart-material-based displays such as polymers and shape-memory alloys[23–28], pneumatic or hydraulic systems[29–33], and unamplified piezoelectric actuators[34–41]. Among these candidates, piezoelectric actuators are especially promising because they combine fast response, high force output, efficient electromechanical transduction, and precise controllability, making them well-suited for tactile displays[42]. However, despite substantial progress, existing tactile displays still face several fundamental limitations that prevent the simultaneous high-fidelity rendering of shape, stiffness, and friction within a single tactile display. First, continuous shape rendering remains rare, and most tactile displays operate in open loop without real-time end-effector feedback. This makes them vulnerable to external disturbances such as variations in fingertip pressure and causes unstable or inaccurate output during active touch[13–21,23–41]. Second, most systems render only a single tactile attribute at a time rather than jointly presenting the multidimensional cues that underlie natural touch perception[1,5,8,28,33]. Third, stiffness rendering remains underdeveloped. The few devices that can modulate stiffness are typically restricted to single-point interaction rather than high-resolution stiffness mapping[23,43–45]. Fourth, many displays rely on perceptual encodings that deviate from natural tactile mappings—for example, using vibration frequency to represent surface height—which increases interpretation difficulty and often requires additional user adaptation or training[13–14,34,40]. Finally, dynamic performance is often insufficient for high-fidelity rendering, particularly in pneumatic systems whose bandwidth remains limited[29,33]. A more detailed review of tactile display technologies is provided in Supplementary Note 1, and a comparative summary is shown in Extended Data Table 1. These limitations motivate the development of a unified tactile rendering platform capable of translating multidimensional physical information into intuitive tactile experiences.

To address these challenges, we present ArrayTac, a programmable tactile platform capable of simultaneously rendering high-fidelity shape, stiffness, and friction cues with continuous tunability. Each actuator unit integrates a three-stage micro-lever amplification mechanism with an end-effector Hall-effect sensor[46], enabling closed-loop displacement control. This end-effector closed-loop architecture enables high-fidelity multidimensional rendering: it compensates for external perturbations during active touch, stabilizes actuator output against fingertip-induced disturbances, and substantially improves rendering speed and accuracy. Coupled with an XYZ sliding exploration platform, the system supports active tactile exploration and enhances spatial perception. In addition, we introduce a stiffness control strategy that emulates the mechanical response of soft materials and enables independent stiffness modulation at each unit. We show that this unified hardware and control framework supports intuitive multidimensional tactile perception, allowing naive participants to directly identify complex three-dimensional shapes and distinguish different stiffness and friction levels through touch alone, without prior training. Building on this capability, we further demonstrate two representative application scenarios for multidimensional tactile interaction: 1) an image-to-touch framework that enables users to perceive object identities, physical attributes, and spatial arrangements from visual input; and 2) a remote palpation framework that enables identification and localization of concealed tumors in a medical-grade breast tumor training phantom over a distance of more than 1,000 km. Together, these contributions establish a tactile platform for simultaneous, high-fidelity, and continuously tunable multidimensional tactile rendering, enabling intuitive tactile communication between humans, machines, and digital or physical environments.

# RESULTS

## Tactile Platform Architecture

ArrayTac is an integrated tactile rendering platform consisting of a closed-loop tactile display array, control electronics, and an XYZ sliding platform. It enables high-fidelity, continuously tunable rendering of shape, stiffness, and friction cues. The overall workflow of the system is shown in Fig. 1a. This system enables tactile information from diverse sources to be rendered and conveyed to the user. The block diagram in Fig. 1b further shows the information-flow architecture of this system. This system-level pipeline enables seamless plug-and-play integration with human-computer interaction (HCI) algorithms and supports tactile interaction at a stable array refresh rate exceeding 500 Hz. The overall system operation is shown in Supplementary Video 1.

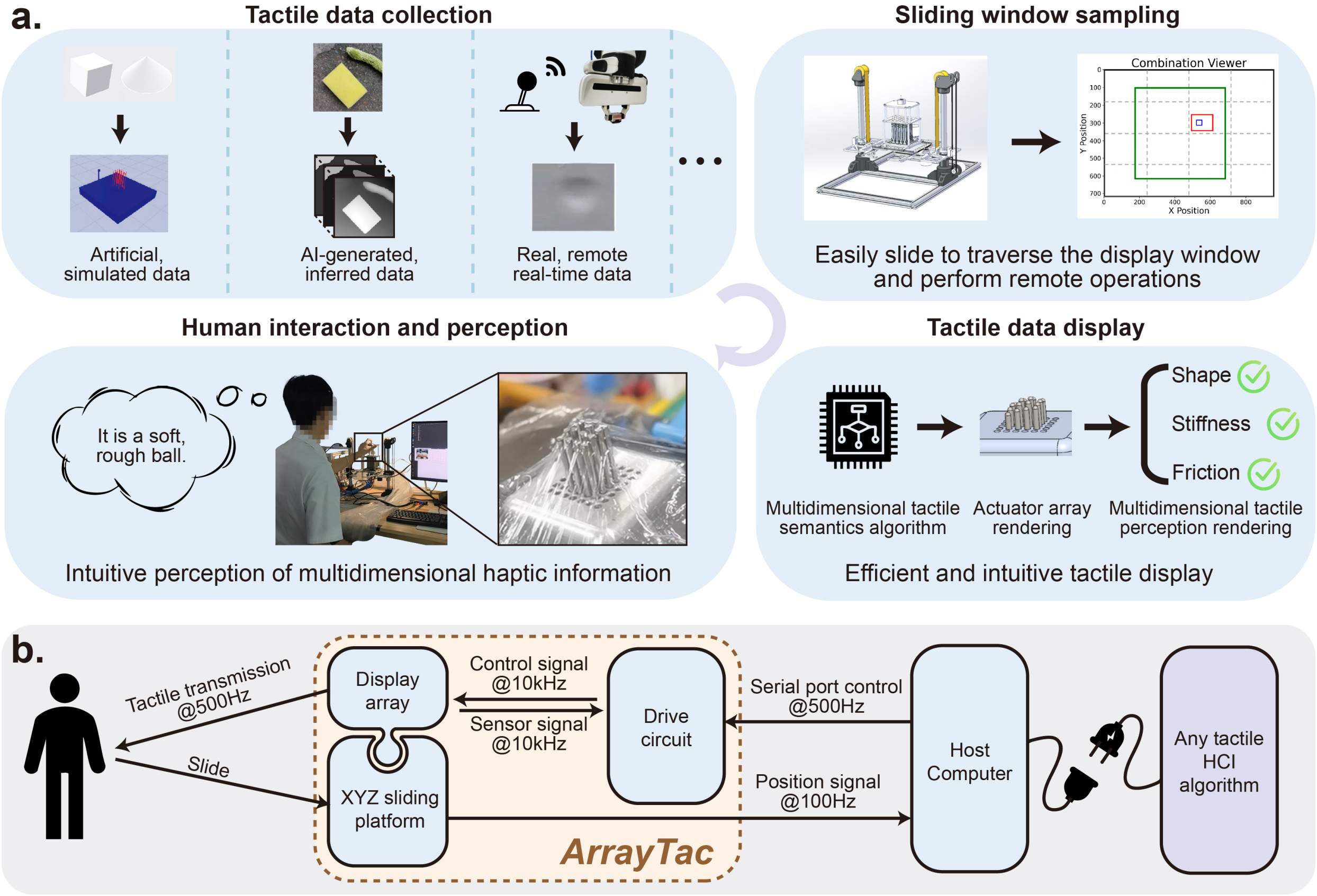

**Fig. 1. Overview of the ArrayTac system. (a)** System workflow and data pipeline from multi-source tactile data collection to haptic rendering. The system first receives multidimensional tactile information from diverse sources—including physical simulations, artificial intelligence (AI)-based inference from images, and real-time remote sensors. These data are then sampled by the user in real time through a "sliding window" to select a region of interest. The sampled information is processed by a multidimensional tactile semantics algorithm that drives the actuator array to render the corresponding shape, stiffness, and friction. Together, these rendered cues provide users with an intuitive multidimensional tactile experience. **(b)** System information-flow diagram. The user controls the end-effector position through the XYZ sliding platform, while the host computer sends the sampled tactile information to the drive circuit via a serial port. The tactile display array then renders multidimensional tactile feedback to the user. This plug-and-play architecture allows the system to integrate with diverse human–computer interaction (HCI) algorithms.

The hardware implementation is summarized in Fig. 2, including an actuator array, a custom drive circuit, an XYZ zero-gravity sliding platform, and the host computer (Fig. 2a). The core of the device is a 4 by 4 array of 16 individually addressable actuator units. Each unit uses a three-stage micro-lever mechanism to amplify the displacement of a piezoelectric ceramic into up to 5 mm end-effector motion (Fig. 2b), while a Hall-effect sensing module provides real-time feedback for closed-loop control. The architecture of the actuator array and the sliding platform are shown in Fig. 2c–e. The unit-level circuit architecture, including the control circuit and power tree, is shown in Fig. 2f. More detailed fabrication and circuit-design considerations are provided in the Methods section, Tactile Platform Design and Fabrication subsection and Supplementary Video 2.

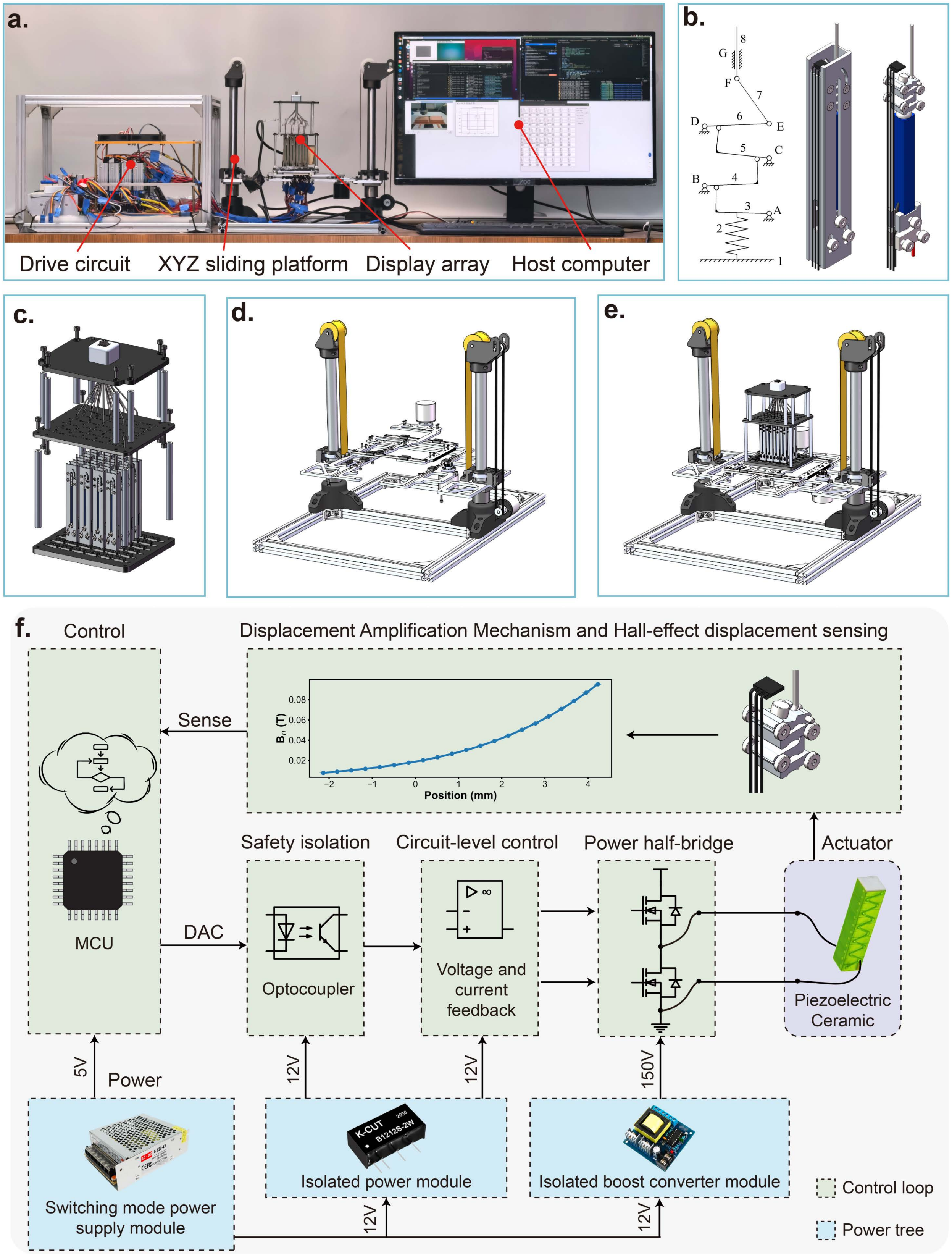

**Fig. 2. System design and schematics. (a)** The complete experimental setup with its main hardware components. **(b)** Left: the schematic of a single actuator unit mechanism. Middle: the external view of a fully assembled actuator unit. Right: an internal perspective view of an actuator unit. **(c)** Exploded view of the actuator array. **(d)** Exploded view of the XYZ sliding stage. **(e)** Assembled view of the interactive platform. **(f)** The circuit schematic for an individual ArrayTac unit, comprising the control circuit and the power tree.

## Multidimensional Tactile Rendering Principles and Performance

For each actuator unit, ArrayTac can simultaneously render shape, stiffness, and friction, while allowing each dimension to be adjusted independently and continuously. Shape is rendered by closed-loop height control. Stiffness is rendered by changing the force–displacement response of each unit. Friction is rendered by vibration cues during sliding. Fig. 3 summarizes the main unit-level control methods and test results.

For shape rendering, the input is a target height for each actuator. Each unit integrates a Hall sensor at the end effector, which measures the motion of the final lever. After calibration, this signal provides end-effector displacement feedback, which is used by a proportional-integral-derivative (PID) controller to track this target height. We first modeled this sensing relation in Ansys EDT and verified its robustness to sensor installation angle changes (Fig. 3a, b). We then calibrated all 16 fabricated units on hardware (Fig. 3c). The closed-loop control allows the controller to reject disturbances from finger contact. Step-response tests showed fast tracking without steady-state error (Fig. 3d). Frequency-response tests showed that closed-loop control increased the effective bandwidth from 15.29 Hz to 123.39 Hz (Fig. 3e, f). This bandwidth supports stable and high-fidelity tactile rendering during active touch.

For stiffness rendering, ArrayTac uses the same feedback structure but changes the control strategy to reproduce the tactile experience of real soft materials. Many soft materials exhibit a two-stage indentation response: they initially resist compression, and then yield after a force threshold is reached. To reproduce this behavior, we use a two-stage controller (Fig. 3g). In the first stage, a PID controller keeps the end effector near the target height and creates a high stiffness response. When the required control effort reaches a threshold set by the normalized stiffness command $k$, a feedforward penalty term changes the drive command. The unit then enters a softer yielding stage whose stiffness is governed by the stiffness command $k$. This matches the response of materials such as high-resilience (HR) foam (Fig. 3h)[47]. We compared different penalty orders in simulation (Fig. 3i): A linear penalty did not capture yielding well. A cubic penalty was too nonlinear and became unstable on hardware. A quadratic penalty gave the best balance between realism and stability. The theoretical analysis in Methods section, Nonlinear Stiffness Rendering Method subsection supports this choice. The experimentally measured displacement–force curves of the display unit are shown in Fig. 3j. We then calibrated the normalized stiffness command against Shore 00 hardness values using a reference indentation setup (Fig. 3k–m). This calibration places the rendered stiffness on a meaningful physical scale.

For friction rendering, previous studies have shown that vibrotactile stimulation applied to the finger during sliding can evoke a friction sensation[48–50]. Accordingly, ArrayTac renders friction by applying programmable vibration during surface exploration. Each actuator adds a sinusoidal vibration to its height command and a normalized friction command $f$ controls the vibration frequency, vibration amplitude, and control gain. Larger commands produce stronger vibration cues. The hardware can generate vibrations from 0 to 512 Hz with amplitudes up to 5 mm. The full control methods, fabrication details, and experimental procedures are provided in Methods section and Supplementary Video 3.

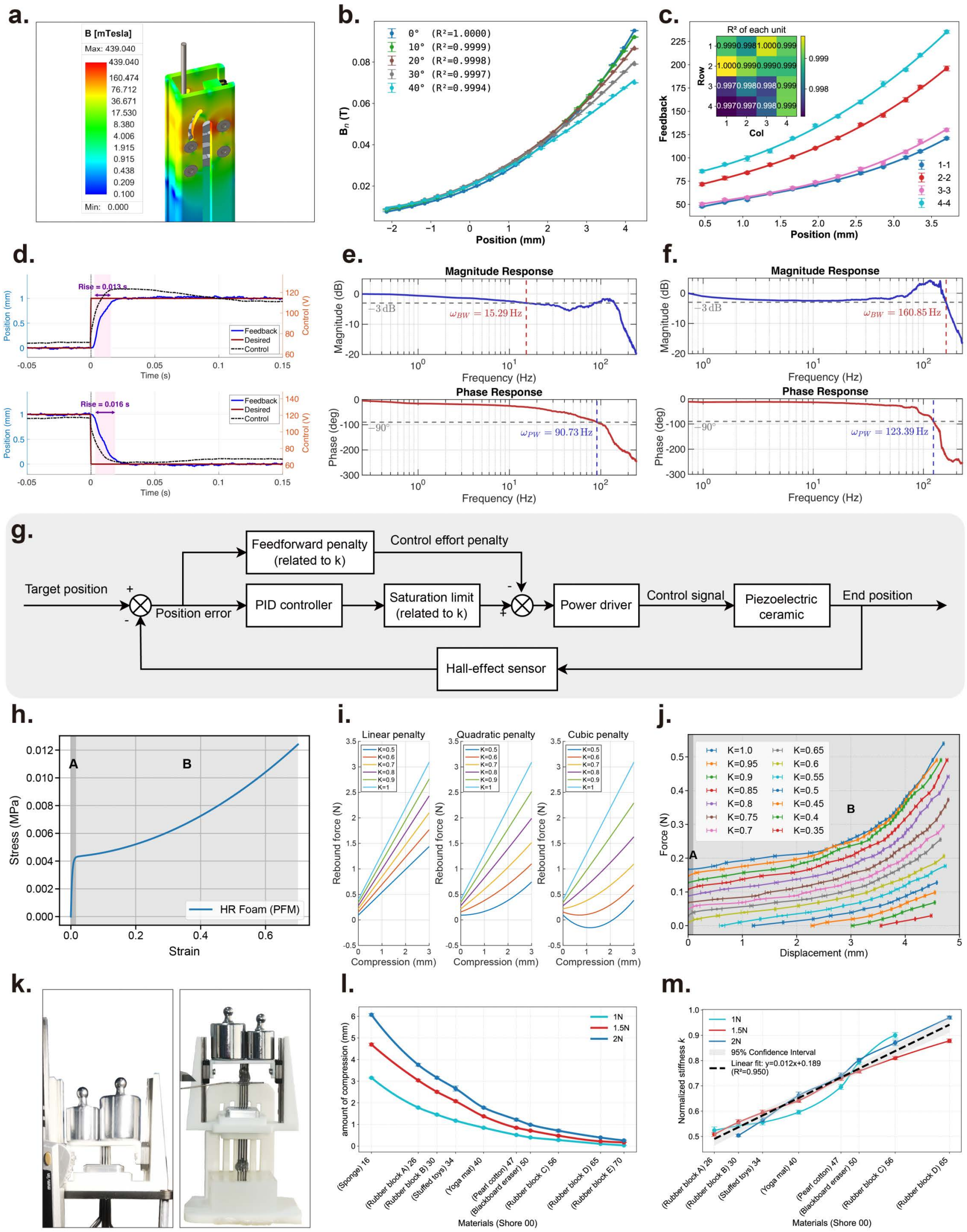

**Fig. 3. Unit-level control method and performance. (a)** Electromagnetic field simulation process performed using Ansys EDT. **(b)** Simulated normal magnetic flux density ($B_n$) detected by the Hall sensor as a function of end-effector position at various sensor installation angles (0° to 40°). The solid lines are cubic polynomial fits to the data, with the coefficient of determination ($R^2$) for each fit shown in the legend. The consistently high values demonstrate the model's robustness against potential assembly misalignments of the sensor ($n = 5$). **(c)** Experimentally measured relationship between Hall-effect sensor feedback and end-effector displacement for each unit. For clarity, only diagonal units are plotted, while the heatmap reports the coefficient of determination ($R^2$) for all 16 units ($n = 3$). For (b) and (c), data points are shown as mean ± s.d., where $n$ represents the number of samples. **(d)** Step response of a single display unit. **(e)** Bode plot of a single display unit under open-loop control. **(f)** Bode plot of a single display unit under PID closed-loop control. For (e) and (f), $\omega_{BW}$ denotes the -3 dB magnitude frequency and $\omega_{PW}$ denotes the -90° phase-lag frequency. **(g)** Block diagram of the stiffness control algorithm. **(h)** Strain-stress curve of HR foam derived from a mathematical model. **(i)** MATLAB-simulated displacement-force curves of the display unit under different penalty orders. **(j)** Experimentally measured displacement-force curves of the display unit under different normalized stiffness parameters ($n = 8$). **(k)** Experimental setup for aligning normalized stiffness with measured Shore 00 hardness. **(l)** Compressive deformation and corresponding Shore 00 hardness of common household objects under different applied normal forces ($n = 5$). **(m)** Relationship between normalized stiffness k and Shore 00 hardness ($n = 5$).

## Psychophysical Evaluation of Multidimensional Tactile Rendering

After validating the rendering mechanisms, we conducted psychophysical experiments to evaluate whether naive users could perceive shape, stiffness, and friction information rendered by ArrayTac. These cues give users geometric, material, and surface information during active touch. We recruited 22 naive participants. All participants reported no tactile impairments and provided informed consent. Fig. 4 summarizes the stimuli, tasks, and perceptual results.

**Shape rendering.** We first tested whether closed-loop height rendering could support direct recognition of 3D geometry. Six geometric models were designed in SolidWorks and processed with V-HACD to generate stable convex hulls for simulation. During each trial, a PyBullet simulator sampled the surface height according to the real-time finger position obtained from the sliding platform. The sampled height map was then sent to the actuator array. To isolate shape perception, the normalized stiffness parameter was fixed at its maximum value, and friction rendering was disabled. The shape study included three tasks:

In **Task 1**, participants explored each rendered shape by touch alone without any other cues or prior training. They then described what they felt. Their responses were scored from 0 to 5 according to descriptive accuracy. A score of 5 indicated perfect identification. A score of 0 indicated complete failure. The zero-shot results in Fig. 4b show that several shapes were intuitive on first use. The hemisphere, cone, and cube reached near-perfect recognition, with median scores of 5. The semi-ellipsoid and bow-shaped platform reached median scores of 4, with participants often reporting curvature, sharp edges, and overall outline. The triangular prism was more difficult, with a median score of 3. This difficulty may reflect the limited spatial resolution of the 4 by 4 actuator array.

In **Task 2**, participants were shown images of six candidate shapes and then identified a randomly rendered shape. Accuracy exceeded 72% for all shape categories (Fig. 4c). Errors mainly occurred between shapes with similar tactile features, which suggests that the display conveyed meaningful geometry.

In **Task 3**, participants received a brief training session. They could see and feel each object before testing. Recognition improved after this short training, and all shapes reached more than 86% accuracy (Fig. 4d). This shape recognition task with prior familiarization was the easiest subtask in our experiment. However, most existing tactile displays require users to be familiarized before recognizing rendered shapes[12,14–15,29–30,40,44,51]. The zero-shot and trained results together show that ArrayTac supports both intuitive and learnable shape perception.

**Stiffness rendering.** We next tested whether the nonlinear stiffness controller could produce intuitive stiffness sensations. We generated five stiffness levels by evenly dividing the normalized stiffness parameter $k$ within the calibrated range from 0.5 to 1.0. This produced a progression from the softest setting to the hardest setting. To isolate stiffness perception, all actuators were fixed at a 3 mm displacement to form a flat surface, and friction rendering was disabled. The stiffness study included two tasks:

In **Task 1**, participants compared two randomly presented stimuli and selected the one that felt stiffer. The pairwise preference heatmap shows strong discrimination (Fig. 4e). Participants distinguished adjacent stiffness levels with accuracy above 86%. For non-adjacent levels, the accuracy reached 100%. This result shows that the five rendered stiffness levels were ordered and perceptually separated.

In **Task 2**, participants performed absolute stiffness identification. After a short familiarization period with the five levels, they classified one randomly presented stimulus. The confusion matrix in Fig. 4f shows that classification accuracy exceeded 74% across all stiffness levels. Most errors occurred between neighboring levels. This result pattern is consistent with gradual stiffness scaling. These results show that the rendered stiffness cues were stable, distinguishable, and memorable.

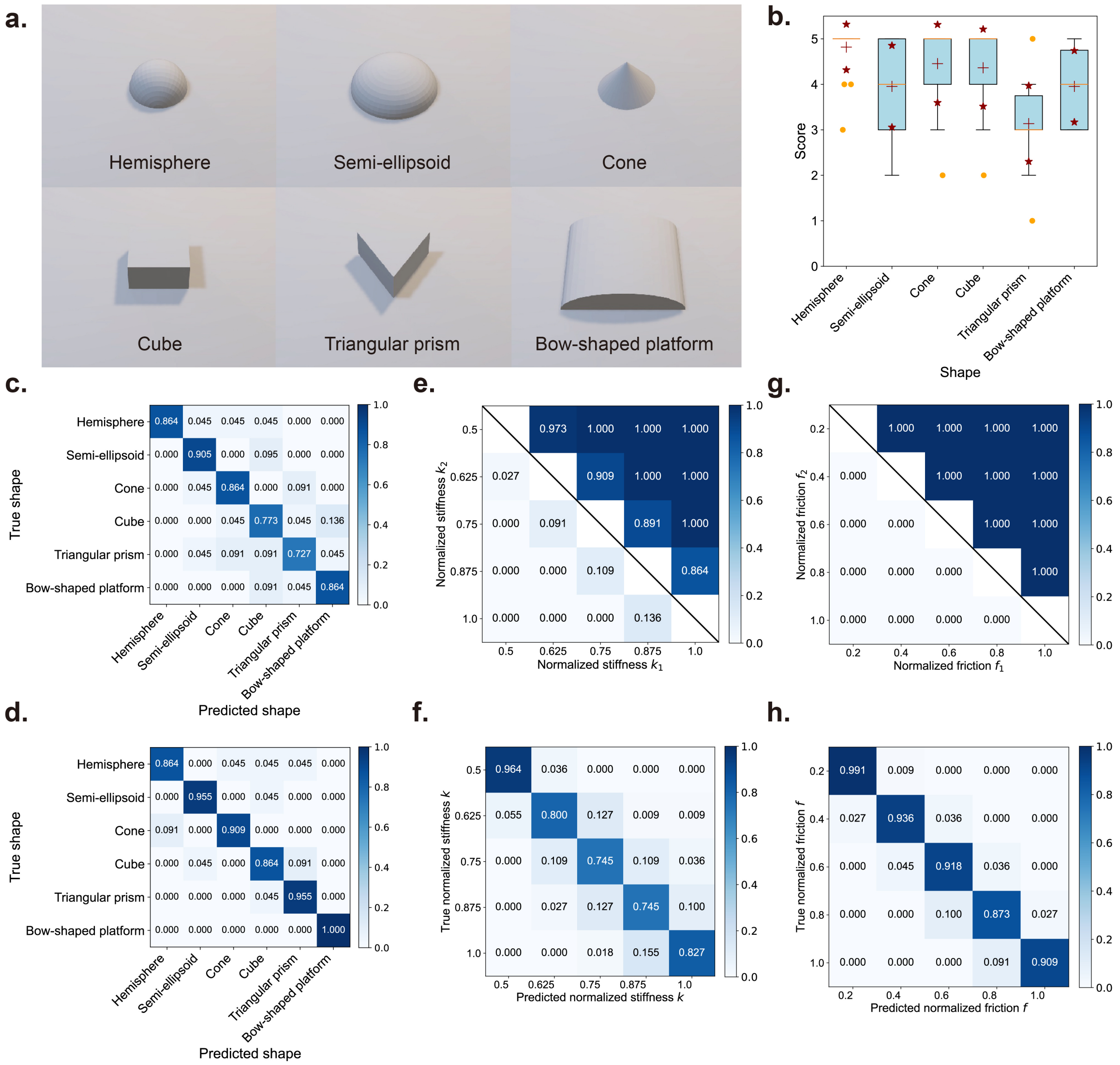

**Fig. 4. Psychophysical experiments on shape, stiffness, and friction perception. (a)** 3D models used in the shape perception experiment. **(b)** Zero-shot performance scores of participants during their first use of the device without any prior training (n = 22). Participants provided open-ended descriptions of each rendered shape, which were evaluated using a predefined 0–5 scoring rubric. A score of 5 indicated exact identification of the rendered shape by name; 4, identification of a closely related shape with only a minor discrepancy; 3, recognition of salient features but omission of important details; 2, provision of limited relevant information together with incorrect information; 1, provision of predominantly irrelevant or incorrect information; and 0, failure to identify any correct feature. In the box plot of (b), the center line indicates the median, the box limits indicate the 25th and 75th percentiles, and the whiskers extend to the maximum and minimum non-outlier values. Outliers are shown as individual orange points. The plus sign indicates the mean, and asterisks indicate one standard deviation above and below the mean. **(c)** Confusion matrix of shape identification when participants were provided only with the list of candidate options (n = 22). **(d)** Confusion matrix of shape identification after participants completed training (n = 22). **(e)** Pairwise preference heatmap for stiffness discrimination, illustrating the proportion of trials in which $k_1$ was judged stiffer than $k_2$ (n = 22; five trials per stimulus pair per participant). **(f)** Confusion matrix of classification across different stiffness levels (n = 22; five trials per level per participant). **(g)** Pairwise preference heatmap for friction discrimination, illustrating the proportion of trials in which $f_1$ was judged rougher than $f_2$ (n = 22; five trials per stimulus pair per participant). **(h)** Confusion matrix of classification across different friction levels (n = 22; five trials per level per participant).

**Friction rendering.** We finally tested the effect of friction rendering. In this experiment, the normalized friction command $f$ was mapped to a frequency range from 0 to 200 Hz. This range was divided into five levels. To isolate friction perception, all actuators were held at a 3 mm displacement, and the stiffness parameter $k$ was fixed at its maximum value. The friction study also included two tasks:

In **Task 1**, participants compared two stimuli and selected the one that felt rougher. They selected the higher friction level with 100% accuracy for every pairwise comparison (Fig. 4g). This result shows that the five rendered friction levels were highly salient.

In **Task 2**, participants classified one randomly presented friction level after familiarization. The confusion matrix shows that accuracy exceeded 87% across all levels (Fig. 4h). These results indicate that each friction level produced a distinct tactile signature.

Together, these experiments demonstrate that ArrayTac enables intuitive perception of multidimensional tactile information without extensive user training. Naive participants could identify 3D shapes and discriminate at least five stiffness and friction levels with high accuracy.

## Tac-Anything Framework and Rendering Experiments

Because ArrayTac enables the high-fidelity, continuously tunable simultaneous rendering of shape, stiffness, and friction, it creates the foundation for a new class of tactile interaction tasks that extend beyond the rendering of individual cues. Building on this capability, we developed Tac-Anything, a novel framework that infers multidimensional tactile information from a single RGB image and translates visual information into tactile experiences through ArrayTac. To our knowledge, this type of direct visual-to-tactile scene understanding, in which object features and spatial information are conveyed through unified multidimensional tactile rendering, has not previously been achieved on a practical tactile platform. We therefore used Tac-Anything to examine whether ArrayTac can support intuitive image-conditioned tactile perception. The complete pipeline is illustrated in Fig. 5a, with a detailed description provided in the Methods section, Tac-Anything Framework subsection.

To evaluate this application scenario, we designed a scene reconstruction experiment. We selected eight common household objects with diverse tactile properties (Fig. 5b) and arranged them into two distinct tabletop scenes (Fig. 5c). Photographs of these scenes were processed by the Tac-Anything framework to generate the multidimensional tactile semantic information, which was sent to ArrayTac. Participants then performed two tasks to evaluate this framework:

**Task 1: Haptic scene sketching.** In this task, participants explored the tactile scene and were asked to sketch their perception of the layout. They used their own notation to annotate the shape, stiffness, and friction properties of each region. We digitized the collected drawings and normalized the stiffness and friction values to a range of $[0,1]$ according to each participant's annotations with the following method: if a participant divided the objects into $n$ levels in stiffness and friction dimensions, each level was assigned values of $0, \frac{1}{n-1}, \frac{2}{n-1}, \dots, 1$. This mapped all ratings to the 0-1 range and ensured comparability across different numbers of levels. Fig. 5d shows the average heatmaps of three tactile semantic dimensions corresponding to the ground truth maps. Quantitatively, the Intersection over Union (IoU) between the sketched shapes and the ground truth was consistently high (Fig. 5e, left plot). The overall mean IoU across all objects and participants was $0.45 \pm 0.15$, with the mean IoU for individual objects ranging from 0.40 to 0.54. These results indicate that the rendered multidimensional tactile information is clear and discriminable, and allows for accurate spatial perception.

**Task 2: Object identification and placement.** This task assessed participants' ability to identify and spatially arrange objects under two distinct conditions. First, participants performed an unrestricted task in which they had to both select objects from eight alternatives and place them on a canvas as precisely as possible. Second, they performed a constrained task where they were informed of the correct object sets and only had to place them. Overall, participants demonstrated high accuracy, with performance significantly improving when participants were given the correct object sets. In the unrestricted condition, the mean placement accuracy was $66.7 \pm 32.5\%$ for Scene 1 and $53.4 \pm 27.1\%$ for Scene 2. Accuracy rose sharply in the constrained condition to $93.9 \pm 19.6\%$ for Scene 1 and $87.5 \pm 24.1\%$ for Scene 2. The results are visualized in the average placement maps (Fig. 5f), and the quantitative results are presented in Fig. 5e (right plot). Notably, the median accuracy of 100% in the constrained tasks indicates that most participants achieved perfect placement. These results confirm that participants could infer object identity and spatial arrangement, with higher accuracy when the candidate object set was constrained.

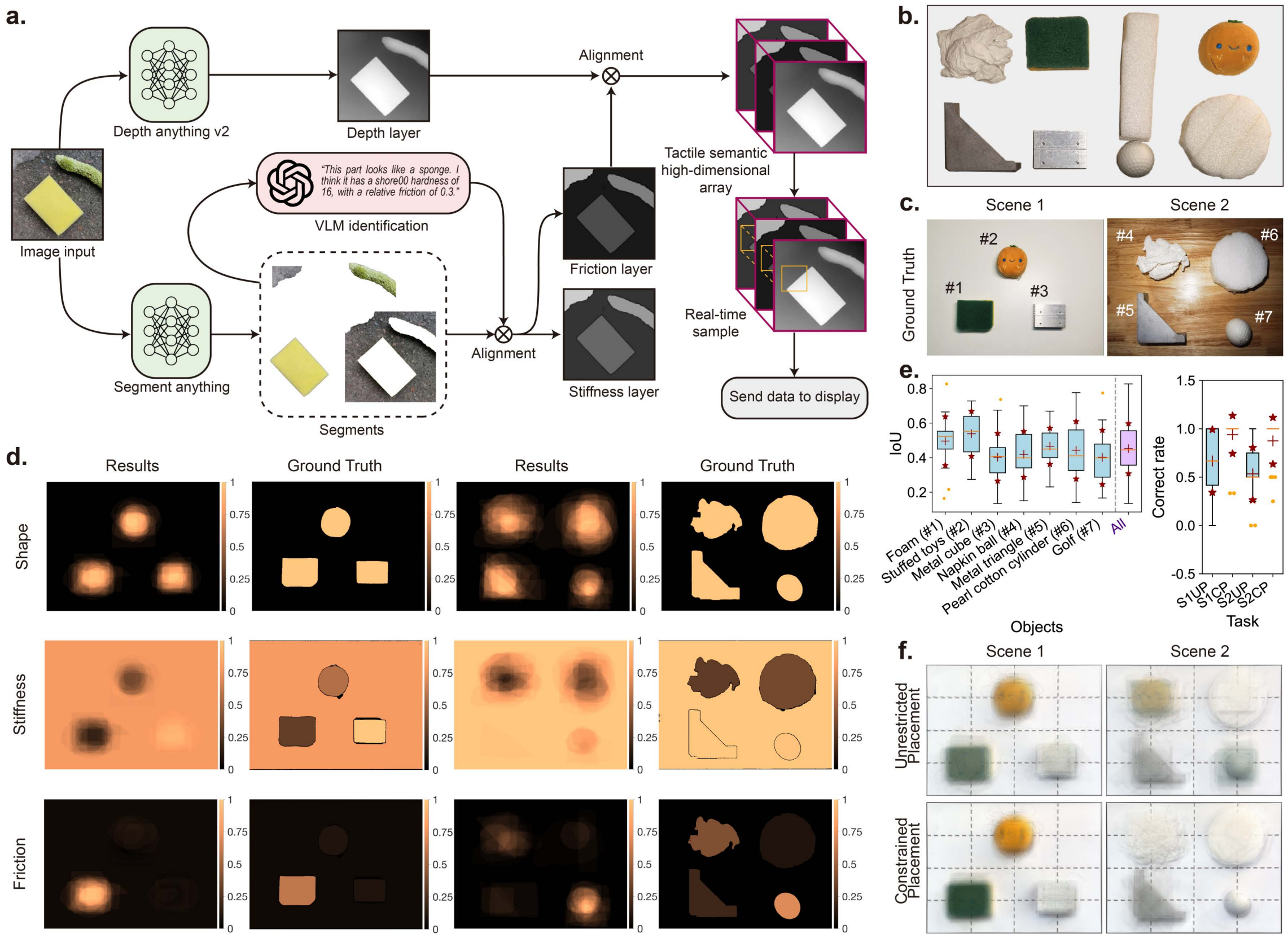

**Fig. 5. The Tac-Anything framework: architecture, experimental validation, and performance. (a)** An overview of the Tac-Anything framework, which extracts and renders multidimensional tactile semantics (shape, stiffness, and friction) from a single RGB image. **(b)** The eight real-world objects with diverse tactile properties used in the user study. **(c)** The two photographic scenes used as the ground truth. **(d)** Average sketching maps of the Haptic Scene Sketching task ($n = 22$). Participant sketches, including their normalized annotations for shape, stiffness, and friction, are compared against the ground truth tactile maps for both scenes. **(e)** Quantitative analysis of participant performance presented as box plots ($n = 22$). The left plot shows the Intersection over Union (IoU) for the Haptic Scene Sketching task for each object and the result for all objects ('All'). The right plot shows the object placement accuracy for the Object Identification and Placement task across four conditions: Scene 1, Unrestricted Placement (S1UP); Scene 1, Constrained Placement (S1CP); Scene 2, Unrestricted Placement (S2UP); and Scene 2, Constrained Placement (S2CP). For the box plots in (e), the center line indicates the median, the box limits indicate the 25th and 75th percentiles, and the whiskers extend to the maximum and minimum non-outlier values. Outliers are shown as individual orange points. The plus sign indicates the mean, and asterisks indicate one standard deviation above and below the mean. **(f)** Average placement maps for the Object Identification and Placement task ($n = 22$), aggregating the final objects and positions chosen by all participants.

## Tele-Touch Experiment for Remote Object Perception

Our tactile rendering platform renders high-fidelity, multidimensional tactile information and can therefore support downstream tasks such as remote palpation. To evaluate this capability, we developed a Tele-Touch system for remote medical palpation. The system's architecture (Fig. 6a) consists of a remote site and a local site connected by a bidirectional flow of data. At the remote site, a robotic arm equipped with a tactile sensor traverses the surface of a tumor phantom, whose internal structure is shown in Fig. 6b, capturing shape information directly and inferring stiffness from force-indentation data. At the local site, the participant teleoperates the robotic arm through the ArrayTac interface, receiving rendered haptic feedback. Fig. 6c shows a live video feed from the remote site, which provides global visual guidance during exploration.

To evaluate the system's effectiveness, we designed two opaque, flesh-like silicone tumor phantoms to emulate clinically relevant palpation conditions (Fig. 6b) for two specific tasks. For the "Tumor Localization" task, the phantom contained tumors of identical size and material. For the "Tumor Severity Ranking" task, the phantom contained tumors of varying size to simulate growth, and of varying hardness (soft silicone vs. hard polylactic acid, PLA) to simulate changes in severity or malignancy. We then conducted these two experiments using the above setup. A detailed implementation of the phantoms and experimental procedures is described in the Methods section, Tele-Touch Pipeline subsection.

**Task 1: Tumor localization.** In this task, participants were asked to determine the number and position of tumors, which were embedded at random locations within the phantom tissue. As shown in Fig. 6d and Fig. 6e, participants identified all tumors and localized them accurately, with tight clustering around the ground truth and consistently low localization errors. The mean localization error was $0.34 \pm 0.15$ cm for the first tumor and $0.41 \pm 0.17$ cm for the second, with an overall mean error of $0.37 \pm 0.16$ cm. These results show that participants could localize concealed targets with sub-centimeter mean error under the tested teleoperation setup.

**Task 2: Tumor severity ranking.** In this task, participants ranked four tumors according to the stiffness and size perceived through the rendered haptic feedback. The pairwise preference heatmap in Fig. 6f shows high values in the upper triangle, indicating reliable severity ranking. To further quantify this, we calculated Kendall's tau ($\tau$) coefficient for each participant's ranking in Fig. 6g. Kendall's tau is a statistic that measures the ordinal association between the participant's ranking and the ground truth, where a value of $\tau = 1$ indicates a perfect match. The distribution of τ coefficients was concentrated near 1.0, with a mean τ of $0.88 \pm 0.21$ (Fig. 6g), confirming that participants could reliably discern tumor properties through the rendered haptic feedback. These results indicate that Tele-Touch can transmit fine tactile details from remote environments, enabling both tumor localization and reliable discrimination of tumor properties in the phantom. The perceptual experiments and representative application demonstrations are summarized in Supplementary Video 4.

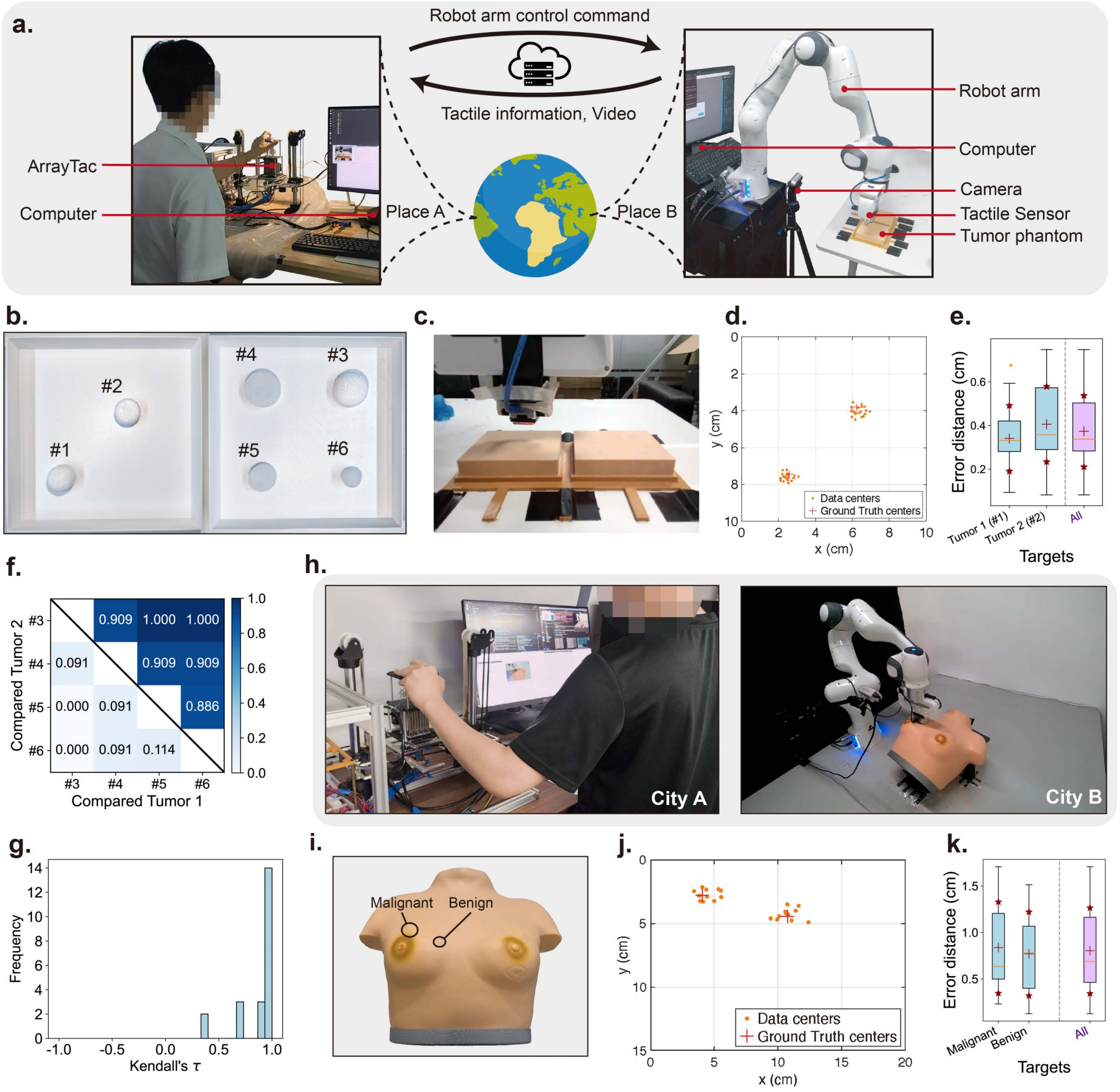

**Fig. 6. Experimental setup and results for the Tele-Touch remote palpation task. (a)** The architecture of the Tele-Touch system. Operators can use the ArrayTac interface locally to manipulate robotic arms located anywhere in the world and perceive remote tactile information. Data streams are transmitted through a cloud server to enable real-time interaction. **(b)** Internal design of the two tumor phantom tissue models. The left phantom was used for the "Tumor Localization" task, while the right phantom was used for the "Tumor Severity Ranking" task. **(c)** Screenshot of the real-time video viewed by participants during the experiment, showing a GelSight tactile sensor mounted on the robotic arm's end effector. **(d)** Visualized results for the tumor localization task ($n = 22$). **(e)** Quantitative analysis of localization error ($n = 22$). The box plot shows the distance between participant-identified centers and the ground truth for two targets and the aggregated data. **(f)** Pairwise preference heatmap for the tumor severity ranking task ($n = 22$). Each cell value represents the percentage of participants who perceived "Compared Tumor 2" (Y-axis) as more severe than "Compared Tumor 1" (X-axis). **(g)** Distribution of Kendall's tau ($\tau$) coefficients for the severity ranking data ($n = 22$). **(h)** Cross-city remote palpation setup. A participant operated ArrayTac at the local site in City A, while a robot arm performed real-time palpation at the remote site in City B, over 1,000 km away, with an end-to-end latency below 0.1 s. **(i)** The medical-grade breast tumor training phantom containing one embedded malignant tumor and one embedded benign tumor. **(j)** Visualized results for the tumor localization task ($n = 11$). **(k)** Quantitative analysis of localization error ($n = 11$). For box plots in (e) and (k), the center line indicates the median, the box limits indicate the 25th and 75th percentiles, and the whiskers extend to the maximum and minimum non-outlier values. Outliers are shown as individual orange points. The plus sign indicates the mean, and asterisks indicate one standard deviation above and below the mean.

## Remote Palpation of a Breast Phantom over 1,000 km

Having validated Tele-Touch in controlled phantom tasks, we next tested its feasibility in a cross-city remote palpation scenario using a medical-grade breast tumor training phantom. The realization of remote palpation carries profound clinical and social implications, potentially enabling patients in remote areas to access expert diagnostic support from specialists at major medical centers. We hope that the high-fidelity tactile rendering capability of ArrayTac can offer a viable technological path toward future remote palpation. Given that palpation is a critical primary screening method for breast tumors, we selected it as our experimental scenario. A planned subset comprising the first 11 participants to register from the full cohort completed the following task. Under the Tele-Touch framework, they operated the ArrayTac system in City A to remotely control a robotic palpation station deployed over 1,000 km away in City B (Fig. 6h). At this remote site, the system interacted with a medical-grade breast tumor training phantom (Fig. 6i), which contained two concealed tumors in the right breast: one malignant and one benign. Despite a straight-line distance exceeding 1,000 km between the two cities, the system maintained a maximum end-to-end latency of under 0.1 s for both control and sensory signals.

Participants had no prior medical expertise or task-specific training. Prior to the task, they were provided with only a basic diagnostic heuristic stating that malignant breast tumors typically exhibit unclear boundaries and uneven surfaces, whereas benign tumors generally have distinct boundaries and smooth surfaces. Relying on the haptic feedback rendered by ArrayTac, participants were asked to determine the number, locations, and types of the hidden tumors and record their findings on paper.

The experimental results were striking: all 11 participants successfully identified the number and types of concealed tumors, achieving a 100% accuracy rate. The spatial localization results from the participants are illustrated in Fig. 6j and Fig. 6k. Quantitative analysis revealed a mean localization error of $0.77 \pm 0.45$ cm for the benign tumor, compared to $0.84 \pm 0.49$ cm for the malignant tumor. The malignant tumor showed a slightly larger localization error than that of the benign tumor, possibly owing to its less distinct boundaries. The cross-city remote palpation demonstration is shown in Supplementary Video 5.

## User Experience and System Performance Evaluation

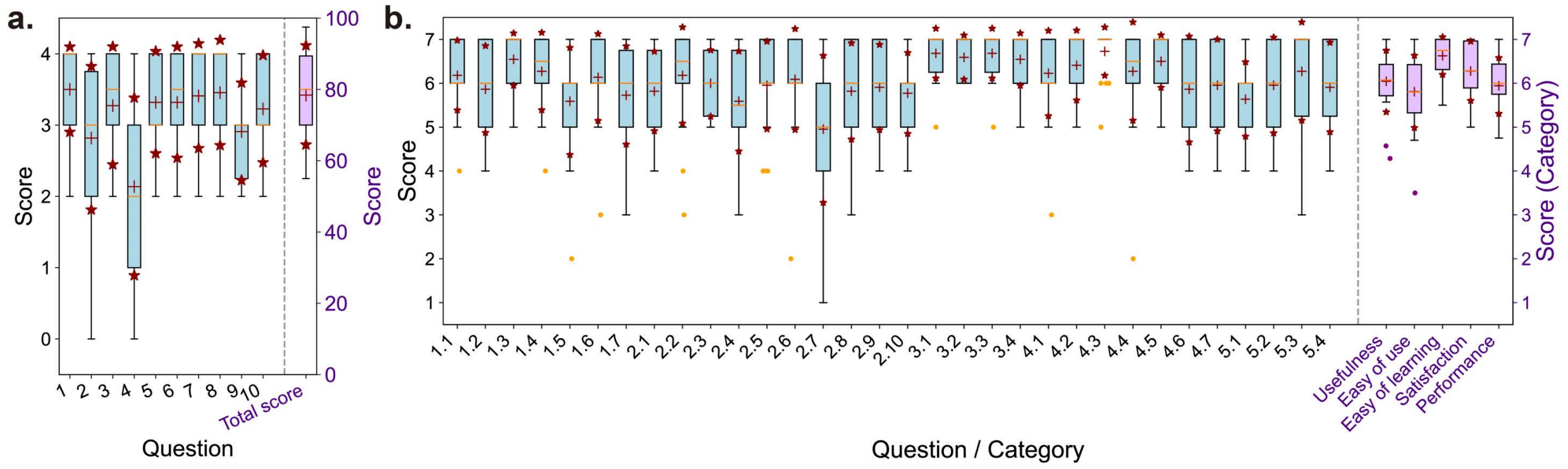

**Extended Data Fig. 1. SUS and USE questionnaire evaluation of ArrayTac. (a)** Distribution of the final calculated SUS scores from all participants, presented as a box plot. The scores are converted from their raw 5-point scale responses to a final 100-point scale using the standard calculation method for standardized benchmarking. For a single question, the scores are converted to 0-4 for easy comparison ($n = 22$). The left y-axis shows the converted item-level SUS scores (0–4), and the right y-axis shows the total SUS score (0–100). **(b)** Distribution of scores for each question in the USE questionnaire, presented as box plots ($n = 22$). For box plots, the center line indicates the median, the box limits indicate the 25th and 75th percentiles, and the whiskers extend to the maximum and minimum non-outlier values. Outliers are shown as individual orange points. The plus sign indicates the mean, and asterisks indicate one standard deviation above and below the mean.

To assess the usability and subjective experience of ArrayTac, 22 participants completed the standard System Usability Scale (SUS) and an adapted USE questionnaire after the main experiments[52–54]. These questionnaires provide a quantitative complement to the perceptual and tele-touch experiments, evaluating whether users found ArrayTac easy to learn, usable, and effective for perceiving the rendered tactile cues.

Extended Data Fig. 1a and Extended Data Fig. 1b show the score distributions of the SUS and adapted USE questionnaires, respectively. Overall, the responses indicate a positive user experience, with participants reporting high perceived usability, learnability, satisfaction, and system performance. Detailed questionnaire design, scoring procedures, and item-level analysis are described in the Methods section, User Questionnaire Survey subsection. The full questionnaire contents and detailed result analysis are provided in Supplementary Table 1, Supplementary Table 2, and Supplementary Note 2.

# DISCUSSION

In this work, we presented ArrayTac, a closed-loop piezoelectric tactile platform for simultaneously rendering shape, stiffness, and friction with continuous tunability. By combining a three-stage micro-lever mechanism with end-effector feedback, the system achieves millimeter-scale displacement, high array refresh rate, and high closed-loop tracking bandwidth, enabling stable and high-fidelity multidimensional rendering during active touch. Psychophysical experiments further showed that naive participants could identify complex shapes and discriminate multiple stiffness and friction levels through touch alone.

We further illustrated the capability of this platform through two representative downstream demonstrations. Tac-Anything combines multimodal foundation models with ArrayTac to translate visual information into tactile perception. Tele-Touch enables cross-city remote palpation over more than 1,000 km, allowing all 11 participants, without prior task-specific training, to correctly identify the number and types of concealed tumors in a medical-grade breast tumor training phantom and localize them with sub-centimeter mean error. Together, these results demonstrate the potential of ArrayTac as a programmable tactile rendering platform for multidimensional haptic interaction, remote tactile communication, and emerging human–machine applications.

This study has several limitations that motivate future work. Although the XYZ sliding platform enables haptic super-resolution and an expanded workspace, the actuator array still has an inherent 4 by 4 spatial resolution. The current system mainly supports fingertip exploration, and uses vibrotactile cues to create a sensation of friction without real tangential forces. Future work could incorporate richer physical models, such as viscoelastic models, to better represent complex soft materials. Despite these limitations, ArrayTac provides a foundation for future tactile communication across human, machine, digital, and physical domains.

# METHODS

## Tactile Platform Design and Fabrication

The tactile platform was engineered as an integrated haptic system. The hardware architecture consists of two main components: the mechanical structure and the drive circuit.

**Mechanical structure.** The complete hardware implementation is shown in Fig. 2a, and the core of the ArrayTac system is a compact 4 by 4 actuator array (Fig. 2c) consisting of 16 independently addressable actuator units. Each unit ($12.4 \times 8.4 \times 76$ mm) integrates a piezoelectric ceramic element, a three-stage micro-lever amplification mechanism, a Hall-effect sensor, and an end-effector steel wire (Fig. 2b). The piezoelectric ceramic (DCS3-050536) generates a displacement of up to 40 $\mu$m, which is amplified 125-fold by the lever system to produce a 5 mm stroke transmitted through a bent steel wire. An axially magnetized NdFeB magnet is mounted on the final lever stage and paired with a Hall sensor (A1324) for closed-loop position feedback. The housing and levers are made of AISI 1045 steel, which provides magnetic shielding and structural stability.

As illustrated in Fig. 2b, the actuator's compact form factor allows dense array integration, while the internal lever system is fabricated via precision wire electrical discharge machining (EDM) to ensure consistent amplification ratios. A fine-threaded adjustment base beneath the piezoelectric element allows accurate height calibration during assembly, compensating for machining and installation tolerances.

The entire actuator array is mounted on a three-degree-of-freedom zero-gravity sliding platform, as shown in Fig. 2d and Fig. 2e. This mechanism expands the tactile workspace to $60 \times 60 \times 185$ mm in the X-Y-Z directions. Orthogonal linear guide rails and low-friction bearings ensure smooth motion, while constant-force springs provide vertical counterbalance, creating a near-weightless interaction experience. The platform's movement is measured through rack-and-pinion and belt-driven rotary encoders, enabling precise, real-time 3D coordinate tracking for synchronized tactile rendering.

**Drive circuit design.** The circuit schematic for an individual ArrayTac unit, illustrated in Fig. 2f, comprises a closed-loop displacement control system and a multi-level power tree. A microcontroller unit (MCU) serves as the core of the control loop. The MCU outputs control signals through two digital-to-analog converter (DAC) channels, which then pass through optocouplers for isolation between the high-voltage and low-voltage domains. The isolated signals serve as reference inputs for circuit-level control. The circuit-level controllers, composed of operational amplifiers, respectively control the upper and lower arms of the power half-bridge. They establish voltage negative feedback for the charging voltage and current negative feedback for the discharging current of the piezoelectric ceramic, thereby achieving high-speed control of the ceramic's voltage. The output of the piezoelectric ceramic, after being amplified by a micro-lever, drives the movement of a permanent magnet, which in turn causes a change in the Hall sensor's reading. After calibration, the displacement of the lever's end can be obtained. This is then used as the feedback input to the MCU to complete the construction of the control loop. The power for the entire system is supplied by a switching power supply. Its 5 V output is used to supply low-voltage systems such as the MCU. Its 12 V output, after being transformed by an isolated power module, is used to provide an isolated power source for the optocoupler and operational amplifiers, ensuring safety and driving capability. This 12 V is also stepped up to 150 V by an isolated boost converter module to serve as the input for the power half-bridge, used to drive the piezoelectric ceramic for full-range actuation. A detailed schematic of the drive circuit is shown in Supplementary Fig. 1.

More detailed information on the fabrication and assembly of the mechanical and electronic components is in the Supplementary Note 3.

## High-Fidelity Shape Rendering System: Design and Validation

Conventional open-loop tactile displays are susceptible to force disturbances from user interaction, which limits their fidelity. To overcome this limitation, we implemented a closed-loop position control system for each of the 16 actuator units. At the core of each unit, a small NdFeB magnet was fixed on the final lever stage, positioned beneath a linear Hall-effect sensor. The sensor measured the change in magnetic flux density, which correlated with the end-effector's vertical displacement, and this analog signal was processed by an onboard microcontroller as the position feedback. Using this real-time feedback, the microcontroller controlled the voltage applied to the piezoelectric element through a Proportional-Integral-Derivative (PID) controller, thereby achieving closed-loop position control. We conducted rigorous experiments to validate the accuracy of the sensor feedback model and quantify the dynamic performance benefits of the closed-loop architecture.

**Validation of the sensor-to-displacement model.** A robust model mapping the Hall sensor's voltage output to physical displacement was established and validated through both simulation and physical experiments. First, the sensor's nonlinear response was modeled in Ansys EDT software. Specifically, we modeled the magnet–sensor architecture in the final lever stage and swept the end-effector position across its operating range to obtain the normal magnetic flux density detected by the Hall sensor. To evaluate tolerance to assembly mismatch, the sensor installation angle was varied from 0° to 40°, and each simulated curve was fitted with the same cubic polynomial form. The simulation of the actuator components (Fig. 3a) revealed that a cubic polynomial provides an exceptionally accurate fit to the data ($R^2 = 0.999$) and is robust to potential assembly variations (Fig. 3b). To confirm these simulation results on the physical hardware, the model was then validated on all 16 fabricated actuators. For hardware calibration, each actuator was driven through a set of known displacement states while the Hall sensor output was synchronously recorded; each unit was fitted independently, and repeated measurements were used to evaluate consistency. For clarity, Fig. 3c plots representative diagonal units, whereas the heatmap reports the fitting accuracy of the full 4 by 4 array. The results were highly consistent with the simulation; applying the same cubic polynomial fit yielded a precise model for every unit, each achieving a coefficient of determination greater than $0.997$ (Fig. 3c). This strong agreement confirmed that our model provides an accurate and reliable method for tracking the end-effector's position.

**Closed-loop control dramatically improves dynamic performance.** To generate realistic and responsive haptic feedback, the display must operate with high speed and stability. We evaluated the actuator performance using step and frequency response tests. During these tests, the actuator tip was placed in contact with a soft silicone pad to approximate the compliant load applied by a fingertip. The results confirmed that closed-loop control is critical for achieving high performance. As shown in Fig. 3d, the step response under overdamped closed-loop control exhibited zero steady-state error, with rise and fall times of 0.013 s and 0.016 s, respectively, ensuring that the device can render sharp features without overshoot. More critically, frequency response analysis revealed a dramatic enhancement in control bandwidth. We characterized the system using a frequency sweep and plotted the resulting Bode response. According to engineering convention, the system bandwidth is defined as the minimum of the -3 dB amplitude cutoff frequency and the -90° phase cutoff frequency. Under the open-loop control (Fig. 3e), the system had a limited bandwidth of only 15.29 Hz, whereas using closed-loop control (Fig. 3f) increased the bandwidth by more than eightfold, to 123.39 Hz. This substantial increase in bandwidth enables the system to effectively compensate for mechanical attenuation and render the complex, high-frequency cues essential for realistic, real-time haptic sensations.

## Nonlinear Stiffness Rendering Method

ArrayTac renders stiffness by programming the force–displacement relationship of each actuator rather than by using a purely linear spring model. The target behavior is the biphasic indentation response observed in many soft materials, such as HR foam, where an initial region with high stiffness is followed by a softer yielding region once the force exceeds a threshold (Fig. 3h)[47]. We therefore implemented a two-stage stiffness controller that combines PID position control with a nonlinear feedforward penalty term (Fig. 3g). The following derivation explains how the controller programs the rendered stiffness and why we selected a quadratic penalty.

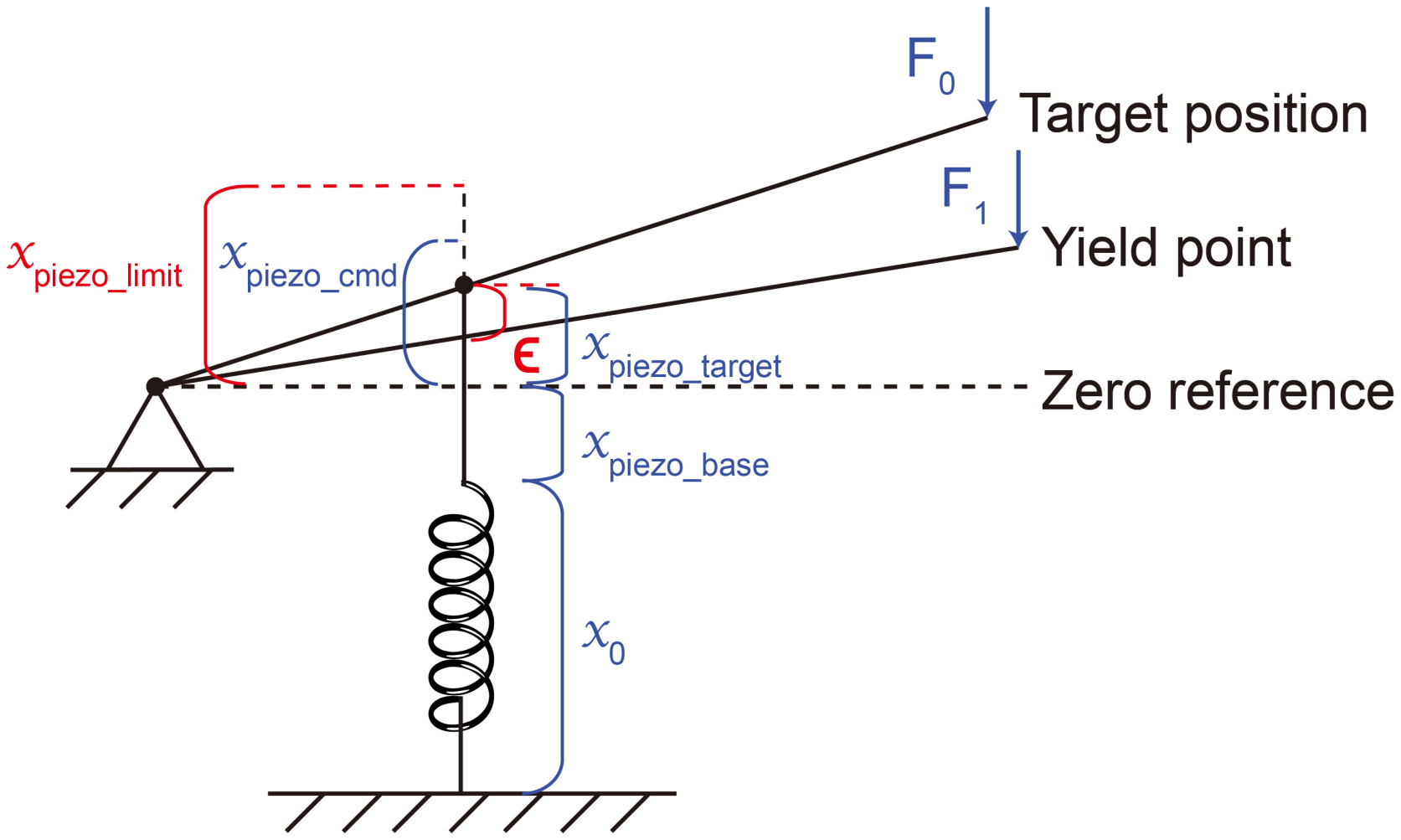

**Extended Data Fig. 2. Physical principle schematic for stiffness rendering.** The piezoelectric element drives a displacement-amplification lever, which converts the piezoelectric displacement into the end-effector displacement. $x_0$ denotes the initial length of the piezoelectric element. The baseline displacement $x_{\text{piezo_base}}$ brings the actuator from its initial position to the mechanical zero reference. The target displacement $x_{\text{piezo_target}}$, command displacement $x_{\text{piezo_cmd}}$, displacement limit $x_{\text{piezo_limit}}$, and displacement error $\epsilon$ of the piezoelectric element define the nonlinear stiffness rendering law. $F_0$ and $F_1$ schematically indicate the normal forces associated with the target position and the yielded position, respectively.

**Actuator model and rendered stiffness.** We modeled each actuator as a piezoelectric element that drives a multi-stage lever for displacement amplification. As illustrated in Extended Data Fig. 2, $x_0$ denotes the initial length of the piezoelectric element, and $x_{\text{piezo_base}}$ denotes the baseline displacement used to bring the actuator to the mechanical zero reference. We denote the piezoelectric force and displacement relative to the mechanical zero reference as $F_{\text{piezo}}$ and $x_{\text{piezo}}$, the end-effector output force and displacement as $F_{\text{actuator}}$ and $x_{\text{actuator}}$, the intrinsic stiffness of the piezoelectric element as $K_{\text{piezo}}$, and the lever displacement amplification gain as $A$ ($A > 1$). The piezoelectric target displacement is $x_{\text{piezo_target}}$, and the corresponding end-effector target displacement is $x_{\text{actuator_target}}$. Under external normal loading, the measured piezoelectric displacement deviates from its target by $\epsilon$, such that $x_{\text{piezo}} = x_{\text{piezo_target}} - \epsilon$. After calibration, the Hall sensor feedback at the end effector provides this displacement error, $\epsilon$.

Assuming an ideal lever, the displacement and force relationships are

$$x_{\text{actuator_target}} = Ax_{\text{piezo_target}} \tag{1}$$

$$x_{\text{actuator}} = Ax_{\text{piezo}} = A\left(x_{\text{piezo_target}} - \epsilon\right) \tag{2}$$

$$F_{\text{actuator}} = \frac{F_{\text{piezo}}}{A} \tag{3}$$

We modeled the piezoelectric element as an elastic actuator. Its force depends on the difference between the commanded no-load displacement relative to the mechanical zero reference, $x_{\text{piezo_cmd}}$, and the actual relative displacement under load, $x_{\text{piezo}}$:

$$F_{\text{piezo}} = K_{\text{piezo}}\left(x_{\text{piezo_cmd}} - x_{\text{piezo}}\right) = K_{\text{piezo}}\left(x_{\text{piezo_cmd}} - x_{\text{piezo_target}} + \epsilon\right) \tag{4}$$

We define the effective stiffness felt by the user as the derivative of the output force with respect to the end-effector deformation, $A\epsilon$:

$$K_{\text{eff}} = \frac{dF_{\text{actuator}}}{d(A\epsilon)} = \frac{1}{A}\frac{dF_{\text{actuator}}}{d\epsilon} \tag{5}$$

Substituting the lever relationships into the piezoelectric force model gives the output force:

$$F_{\text{actuator}} = \frac{K_{\text{piezo}}}{A}\left(x_{\text{piezo_cmd}} - x_{\text{piezo_target}} + \epsilon\right) \tag{6}$$

Because the controller updates $x_{\text{piezo_cmd}}$ according to the measured error $\epsilon$, differentiating equation (6) with respect to $\epsilon$ gives

$$\frac{dF_{\text{actuator}}}{d\epsilon} = \frac{K_{\text{piezo}}}{A}\left(\frac{dx_{\text{piezo_cmd}}}{d\epsilon} + 1\right) \tag{7}$$

Substituting equation (7) into equation (5) gives

$$K_{\text{eff}} = \frac{K_{\text{piezo}}}{A^2}\left(\frac{dx_{\text{piezo_cmd}}}{d\epsilon} + 1\right) \tag{8}$$

Equation (8) separates the rendered stiffness into a passive term, $\frac{K_{\text{piezo}}}{A^2}$, and an active term determined by the controller, $\frac{K_{\text{piezo}}}{A^2}\frac{dx_{\text{piezo_cmd}}}{d\epsilon}$. Thus, by specifying how the commanded displacement changes with the error $\epsilon$, the system can program the effective stiffness felt at the end effector.

**Nonlinear controller for two-stage stiffness rendering.** Based on this model, we implemented a controller that reproduces the two-stage indentation profile of soft materials. First, the controller drives the actuator to its mechanical zero reference using the baseline piezoelectric displacement $x_{\text{piezo_base}}$. The controller then enters a stage dominated by PID position control, where it maintains the target height and keeps $\epsilon \approx 0$ as the external normal force increases. This stage corresponds to the initial region of the HR-foam response, where the material exhibits high stiffness. Once the required control effort reaches the saturation threshold set by the stiffness command $k$, the system can no longer maintain zero error. The feedforward penalty then uses the resulting $\epsilon$ to reduce the effective drive command, producing a yielding response corresponding to the softer region of the material curve.

We set the saturation threshold in the voltage domain as

$$U_{\text{limit}} = \frac{k+4}{5}\left(U_{\max} - U_{\text{base}}\right) + U_{\text{base}} \tag{9}$$

where $k \in [0,1]$ is the normalized stiffness command, $U_{\max}$ is the maximum driving voltage, and $U_{\text{base}}$ is the voltage corresponding to the zero reference. Let $\beta$ denote the coefficient that converts voltage to piezoelectric displacement. The baseline and threshold displacements are then

$$x_{\text{piezo_base}} = \beta U_{\text{base}},\ x_{\text{piezo_limit}} = \beta\left(U_{\text{limit}} - U_{\text{base}}\right) \tag{10}$$

where $x_{\text{piezo_limit}}$ represents the maximum relative piezoelectric displacement available before the feedforward penalty is applied. Below the saturation threshold, PID control maintains the end effector at the target position. After the controller reaches this threshold, the nonlinear penalty modifies the output voltage according to

$$U_{\text{output}} = \left(U_{\text{limit}} - U_{\text{base}}\right)\left(1 - (1-k)\frac{\epsilon}{x_{\text{piezo_limit}}}\right)^n + U_{\text{base}} \tag{11}$$

where $n$ controls the nonlinearity of the yielding response. Because $x_{\text{piezo_cmd}}$ is defined relative to the mechanical zero reference, the same command can be written in displacement form as

$$x_{\text{piezo_cmd}}(\epsilon) = x_{\text{piezo_limit}}\left(1 - \frac{1-k}{x_{\text{piezo_limit}}}\epsilon\right)^n \tag{12}$$

This formulation gives $k$ two roles: it sets the saturation threshold for the transition into yielding, and it controls how strongly the command decreases after yielding.

**Selection of the penalty exponent.** To avoid unphysical rendering behavior, the nonlinear control law should satisfy two constraints. First, the rendered effective stiffness should remain non-negative to avoid mechanically unstable negative-stiffness behaviour:

$$K_{\text{eff}} \geq 0 \tag{13}$$

Second, because many soft materials stiffen under increasing deformation, the effective stiffness should also be non-decreasing with deformation:

$$\frac{dK_{\text{eff}}}{d\epsilon} \geq 0 \tag{14}$$

From equation (8), these two requirements translate into the following conditions on $x_{\text{piezo_cmd}}(\epsilon)$:

$$\frac{dx_{\text{piezo_cmd}}}{d\epsilon} \geq -1 \tag{15}$$

and

$$\frac{d^2x_{\text{piezo_cmd}}}{d\epsilon^2} \geq 0 \tag{16}$$

Differentiating the penalty law in equation (12) gives

$$\frac{dx_{\text{piezo_cmd}}}{d\epsilon} = -n(1-k)\left(1 - \frac{1-k}{x_{\text{piezo_limit}}}\epsilon\right)^{n-1} \tag{17}$$

and

$$\frac{d^2x_{\text{piezo_cmd}}}{d\epsilon^2} = n(n-1)\frac{(1-k)^2}{x_{\text{piezo_limit}}}\left(1 - \frac{1-k}{x_{\text{piezo_limit}}}\epsilon\right)^{n-2} \tag{18}$$

We considered three candidate penalty exponents, $n = 1, 2, 3$. For a linear penalty $(n = 1)$, equation (8) gives:

$$K_{\text{eff}} = k\frac{K_{\text{piezo}}}{A^2}, \quad \frac{d^2x_{\text{piezo_cmd}}}{d\epsilon^2} = 0 \tag{19}$$

Although this case allows $k$ to tune the apparent stiffness, the resulting stiffness does not vary with deformation and therefore fails to meet the stiffening requirement.

For a quadratic penalty $(n = 2)$, differentiating equation (12) gives

$$\frac{dx_{\text{piezo_cmd}}}{d\epsilon} = -2(1-k)\left(1 - \frac{1-k}{x_{\text{piezo_limit}}}\epsilon\right) \geq 2k - 2 \tag{20}$$

and

$$\frac{d^2x_{\text{piezo_cmd}}}{d\epsilon^2} = \frac{2(1-k)^2}{x_{\text{piezo_limit}}} \geq 0 \tag{21}$$

Thus, when $k \geq 0.5$, the effective stiffness remains non-negative over the operating range and is strictly positive for $k > 0.5$. The second derivative is positive for $k < 1$ and becomes zero at $k = 1$.

For a cubic penalty $(n = 3)$:

$$\frac{dx_{\text{piezo_cmd}}}{d\epsilon} = -3(1-k)\left(1 - \frac{1-k}{x_{\text{piezo_limit}}}\epsilon\right)^2 \geq 3k - 3 \tag{22}$$

Although this case can also maintain positive effective stiffness when $k \geq 0.67$, it narrows the usable range of $k$. In hardware tests, the stronger nonlinearity also produced unstable haptic feedback. Therefore, we selected the quadratic penalty $(n = 2)$ as the simplest nonlinear term that satisfies these constraints while maintaining a stable operating range.

**Experimental validation and calibration to Shore 00 hardness.** We validated the nonlinear stiffness controller in both simulation and hardware. In MATLAB simulations, we compared linear, quadratic, and cubic penalty terms and found that the nonlinear penalties better reproduced biphasic force–displacement curves (Fig. 3i). In hardware tests, the cubic penalty was unstable, whereas the quadratic penalty produced a stable and realistic yielding response. We therefore used $n = 2$ for all subsequent stiffness rendering experiments. The measured force–displacement curves under different normalized stiffness values confirmed that the controller reproduced two-stage responses similar to those of HR foam (Fig. 3j). The parameter $k$ modulated both the initial resistance and the response after yielding, enabling continuous tuning of nonlinear stiffness.

Finally, we calibrated the normalized parameter $k$ against Shore 00 hardness values to relate the rendered stiffness to a standard material metric. We used a reference indentation setup whose contact geometry matched that of the tactile display unit (Fig. 3k). For each reference object, we measured the compressive deformation under three fixed normal forces, $F_n = 1, 1.5, 2$ N (Fig. 3l). For each force level, we adjusted $k$ in ArrayTac until the rendered deformation matched the deformation measured from the reference object. This procedure produced a near-linear mapping between the calibrated $k$ values and Shore 00 hardness (Fig. 3m). This mapping allowed us to interpret the stiffness levels used in the psychophysical experiments on a physical material scale.

## Friction Rendering Method

During sliding contact, friction perception depends not only on tangential resistance but also on high-frequency vibrations generated at the contact interface[48–50]. Accordingly, ArrayTac renders friction by superimposing a sinusoidal vibration on the baseline displacement command for shape rendering. For each actuator, the normalized friction command $f \in [0,1]$ modulates the vibration frequency, amplitude, and control gain. Larger $f$ values produce stronger vibrotactile cues while preserving closed-loop shape rendering. The actuator hardware supported vibration frequencies from 0 to 512 Hz and amplitudes up to 5 mm. In the psychophysical friction experiments, $f$ was mapped to 0–200 Hz and divided into five levels; all actuators were held at a flat 3 mm height, and stiffness was fixed at $k = 1$ to isolate friction perception. For remote palpation experiments, friction rendering was disabled by setting $f = 0$.

## Tac-Anything Framework

To enable participants to perceive tactile properties from a single 2D image, we developed the Tac-Anything framework. Although previous studies have explored image-to-touch conversion, these approaches were limited to transforming scenes into predefined symbolic patterns. Such representations often fail to align with natural haptic semantics and sensory intuition, and they typically lack real-time rendering capability[55]. In contrast, Tac-Anything extracts multidimensional tactile information, including shape, stiffness, and friction, from a single photograph and renders it on ArrayTac in real time. The framework's architecture is illustrated in Fig. 5a. The workflow begins with a single RGB image as input, which is processed simultaneously through two parallel deep learning pathways. In the first pathway, the system reconstructs the surface topography. The image is processed by the Depth-Anything-V2 model[56] to perform monocular depth estimation. This step generates a dense depth map (the "Depth layer") that represents the scene's 3D shape. Concurrently, the second pathway determines material properties. The image is sent to the Segment-Anything model[57], which identifies and isolates the main objects. This process generates distinct "segments" and their corresponding masks. Each isolated segment is then analyzed by a Vision Language Model (VLM). The VLM leverages its world knowledge to infer quantitative tactile properties, such as "estimated Shore 00 hardness of 16" and "estimated relative friction of 0.3". This inferred data is aligned with the object masks to create the "Stiffness" and "Friction" layers. Finally, the three generated layers—Depth, Stiffness, and Friction—are spatially aligned and stacked. This creates a unified "tactile semantic high-dimensional array". The array is then sampled in real time according to the user's interaction, and the resulting tactile information is streamed to ArrayTac, allowing participants to perceive the tactile properties from the input image.

## Tele-Touch Pipeline

We constructed a Tele-Touch system to validate ArrayTac's performance in a remote touch scenario. The architecture of this system is illustrated in Fig. 6a. The system is divided into two sites: a local site (Site A) and a remote site (Site B), with communication established via a cloud-based TCP server. Site A consists of our haptic display and the user, while Site B comprises a Franka robotic arm equipped with a GelSight visuotactile sensor and the target materials for palpation.

During operation, the user at Site A manipulates the tactile display, moving it in XYZ space. This movement is translated into commands that teleoperate the robotic arm at Site B, enabling it to explore and press the target materials in three dimensions. Simultaneously, the GelSight sensor at the remote site captures high-resolution tactile information upon contact. These data are processed and transmitted back to the display at Site A for real-time haptic rendering. The feedback includes shape information, which is read directly from the GelSight sensor, and stiffness information, which is calculated as the ratio of the applied normal force to the measured compression depth. As friction is considered less critical for this palpation task, its value was preset to 0.

## Human participant experiments

We recruited 22 self-reported healthy adults (20 male and 2 female; age range, 19–26 years; mean age, 22.05 years; s.d., 1.86 years). All participants reported normal hand motor and tactile function, met the prespecified eligibility criteria, and were first-time users of ArrayTac before participation. No prior expertise in medicine, robotics, haptics or human–computer interaction was required. All participants completed the main task modules in the same fixed order, while stimulus identities and presentation sequences were randomized within the relevant discrimination and classification tasks. All 22 participants completed the shape, stiffness, friction, Tac-Anything, controlled Tele-Touch and questionnaire experiments. A planned subset consisting of the first 11 participants to register from the same cohort additionally completed the cross-city breast-phantom demonstration. No participants or trials were excluded from the reported analyses.

## User Questionnaire Survey

To evaluate the usability and subjective user experience of ArrayTac, we administered two questionnaires after participants completed the main experimental tasks. All 22 participants completed the standard System Usability Scale (SUS) and an adapted USE questionnaire[52,53]. The SUS questionnaire was included to provide a standardized usability benchmark, while the USE questionnaire was used to capture broader aspects of user experience that were more closely related to the tactile rendering tasks, including perceived usefulness, ease of use, learnability, satisfaction, and system performance. All results are reported as box plots in Extended Data Fig. 1.

SUS scores were calculated using the standard scoring procedure[53]. For odd-numbered positive items, the item contribution was computed as the selected scale position minus 1. For even-numbered negative items, the item contribution was computed as 5 minus the selected scale position. This converts each SUS item to a 0–4 contribution score. The ten converted item scores were then summed and multiplied by 2.5 to obtain the final SUS score on a 0–100 scale. The final SUS score was used as a standardized measure of perceived usability, with higher scores indicating better usability.

The SUS results in Extended Data Fig. 1a showed that ArrayTac achieved a final score of 78.41 ± 13.92, falling just below the A- threshold according to published SUS benchmarks[54]. Most SUS items received high converted scores, indicating that participants generally found the system easy to learn and were willing to use it frequently. The only relatively lower item was related to whether users could operate the system without technical support, suggesting that a brief introduction or initial guidance may still be helpful before fully independent use.

The USE questionnaire was adapted to the experimental context by removing items that were not applicable to ArrayTac and by adding a Performance dimension. This additional dimension was designed to evaluate whether participants perceived the rendered tactile cues as effective, reliable, and helpful during the experimental tasks. All USE items were rated on a 7-point Likert scale, where 1 indicated "strongly disagree" and 7 indicated "strongly agree." Dimension-level scores were computed by averaging the items within each dimension for each participant.

The adapted USE questionnaire results shown in Extended Data Fig. 1b further supported this positive subjective evaluation. Thirty of the thirty-two items achieved median scores of 6 or higher, indicating strong agreement across most aspects of user experience. Among the five evaluated dimensions, Learnability received the highest average score, suggesting that participants could quickly understand how to interact with ArrayTac. Satisfaction and Usefulness were also rated highly, indicating that participants perceived the tactile feedback as meaningful and valuable during the tasks. The Performance dimension received a similarly positive score, suggesting that users regarded the rendered tactile cues as effective for supporting perception. The slightly lower response to the item "I can use it without written instructions" was consistent with the SUS result, again indicating that short instructions can improve user confidence.

The questionnaire results suggest that ArrayTac was perceived as highly usable, easy to learn, and effective for conveying rendered tactile cues. Together with the high perceptual accuracy observed in the previous tasks, these subjective evaluations indicate that participants experienced the system as intuitive and acceptable for active tactile interaction.

**Extended Data Table 1. Comparison of representative tactile displays.** The table summarizes whether each system supports shape, stiffness, and friction rendering, whether multiple tactile dimensions can be rendered simultaneously on the same interface, and whether the displayed cue is mapped through a natural physical instantiation. ArrayTac is distinguished by end-effector closed-loop control and the simultaneous, continuously tunable rendering of shape, stiffness, and friction on a unified fingertip tactile platform. Note. "Continuous shape rendering" is marked as Yes only when the rendered surface height itself varies continuously across levels; binary states or other indirect representations are marked as No. "Simultaneous rendering" is marked positive only when two or more tactile dimensions are rendered at the same time on the same display during a single interaction. "Natural cue mapping" is marked as Yes only when the target tactile cue is directly physically instantiated at the interface; indirect encoding is marked as No. Underlined values were not directly reported and were instead estimated from the original papers. NR, not reported; N/A, not applicable.

| Work | Actuation strategy | Display format and resolution | Closed-loop end-effector feedback | Continuous shape rendering | Stiffness rendering | Friction rendering | Simultaneous rendering | Natural cue mapping | Bandwidth | Refresh rate | Max stroke |
|---|---|---|---|---|---|---|---|---|---|---|---|
| FiDTouch *(12)* | Mechanical | Fingertip, single-point contact | No | No (only one point) | No | No | No | No | NR | NR | 8 mm |
| Shapeshift *(11)* | Mechanical | Tabletop, 12 × 24 pin array, 7 mm pitch | No | **Yes** | No | No | No | **Yes** | NR | 50 Hz | 50 mm |
| P. Kammermeier, et al. *(13)* | Electromagnetic | Fingertip, 6 × 6 pin array, 3 mm pitch | No | No (binary) | No | No | No | **Yes** | >100 Hz | NR | 1.6 mm |
| HUGO *(14)* | Electromagnetic | Fingertip, 5 × 5 pin array, 2 mm pitch | No | **Yes (need active scanning)** | No | **Yes** | **Shape, Friction** | No | NR | 110 Hz | 3 mm |
| M. M. H. Saikot, et al. *(15)* | Electromagnetic | Fingertip, 2 × 3 pin array, adjustable 2.3-3.25 mm pitch | No | No (binary) | No | No | No | **Yes** | NR | >10 Hz | 2 mm |
| D. Chen, et al. *(16)* | Electromagnetic | Fingertip, 24 pluggable 2 × 3 pin array modules, 2.5 mm pitch | No | No (binary) | No | No | No | **Yes** | NR | 2.7 Hz | 0.5 mm |
| T. Iwamoto, et al. *(17)* | Ultrasound | Fingertip, scanned focal point, 1 mm focal diameter | No | No (scanned pressure pattern) | No | No | No | No | 1,000 Hz | NR | N/A |
| T. Hoshi, et al. *(18)* | Ultrasound | Palm, scanned focal point, 20 mm focal diameter | No | No (scanned pressure pattern) | No | No | No | No | 1,000 Hz | NR | N/A |
| P.-H. Chu, et al. *(21)* | Electrical Stimulation | Fingertip, 4 electrodes | No | No | No | No | No | No | NR | NR | N/A |
| S. Kim, et al. *(23)* | Magnetorheological | Fingertip, single-cell cube, 20 mm cube | No | No | **Yes** | No | No | **Yes** | NR | <3 Hz | NR |
| I. Hwang, et al. *(24)* | Photothermal pneumatic | Tabletop, 6 × 6 morphable film array, 5 mm pitch | No | **Yes** | **Yes** | No | **Shape, Stiffness** | **Yes** | NR | <0.5 Hz | 1.42 mm |
| P. S. Wellman, et al. *(25)* | Shape memory alloy | Fingertip, 10-pin line array, 2 mm pitch | **Yes** | **Yes** | No | No | No | **Yes** | 40 Hz | NR | 3 mm |
| S.-Y. Jang, et al. *(26)* | Electrohydraulic | Tabletop, 5 × 5 electrode array, unit size not reported | No | **Yes** | No | **Yes** | **Shape, Friction** | No | NR | NR | 2.5 mm |
| C. Tian, et al. *(27)* | Electroactive polymer | Fingertip, 2 × 3 cell, 3.75 mm pitch | No | No (binary) | No | No | No | **Yes** | NR | NR | 0.586 mm |
| VoxeLite *(28)* | Electroadhesive | Fingertip, up to 8 × 8 node array, 1 mm pitch | No | No | No | No | No | No | NR | NR | 0.081 mm |
| Y. Ujitoko, et al. *(29)* | Pneumatic | fingertip, 128 pins (12×10+8 staggered), 1.4 mm pitch | No | No (binary) | No | No | No | **Yes** | NR | 50 Hz | 10 mm |
| J. Jeong, et al. *(30)* | Hydraulic | Fingertip, single-point contact | No | No (force display) | No | No | **Pressure, Temperature** | **Yes** | NR | NR | 10.1 mm |
| B. Shan, et al. *(31)* | Microfluidic | Fingertip, 5 × 5 microfluidic actuator array, 1.25 mm pitch | No | **Yes** | No | No | No | **Yes** | NR | <2 Hz | 0.145 mm |
| HDAM *(32)* | Piezoelectric, Hydraulic | Fingertip, 3 × 3 array, 3 mm pitch | No | **Yes** | No | No | No | **Yes** | NR | <140 Hz | 0.1 mm |
| R. H. Heisser, et al. *(33)* | Microcombustion | Fingertip, 10 × 10 dot array, 2.5 mm pitch | No | No (binary) | No | No | No | **Yes** | NR | 0.04 Hz | 1 mm |
| K.-U. Kyung, et al. *(34)* | Piezoelectric | Fingertip, 6 × 5 pin array, 1.8 mm pitch | No | No (binary) | No | No | No | **Yes** | 360 Hz | NR | 0.7 mm |
| MIMIZU *(36)* | Piezoelectric | Tabletop, 32 × 24 pin matrix, 3 mm pitch | No | No (binary) | No | No | No | **Yes** | NR | NR | 0.7 mm |
| G.-H. Yang, et al. *(37)* | Piezoelectric | Fingertip, 6 × 5 pin array, 1.8 mm pitch | No | No (binary) | No | No | **Shape, Temperature** | **Yes** | NR | NR | 0.7 mm |
| STReSS *(38)* | Piezoelectric | Fingertip, 10 × 10 array, 1 mm pitch | No | No | No | No | No | No | NR | 700 Hz | 0.05 mm |
| Y. Ikei, et al. *(40)* | Piezoelectric | Fingertip, 5 × 10 pin array, 2 mm pitch | No | **Yes** | No | No | No | No | NR | NR | 0.057 mm |
| G. Frediani, F. Carpi *(44)* | Pneumatic | Fingertip, single deformable chamber, 12 mm chamber diameter | No | No | **Yes** | No | No | **Yes** | 3 Hz | 20 Hz | 6 mm |
| **ArrayTac (ours)** | **Piezoelectric** | **Fingertip, 4 × 4 actuator array, 1.7 mm pitch** | **Yes** | **Yes** | **Yes** | **Yes** | **Shape, Stiffness, Friction** | **Yes** | **123 Hz** | **>500 Hz** | **5 mm** |